\documentclass{cvmp_short}

\usepackage{graphicx}
\usepackage{booktabs}
\usepackage{graphicx}
\usepackage{url}
\usepackage{times}
\usepackage{epsfig}
\usepackage{graphicx}
\usepackage{amsmath}
\usepackage{amssymb}
\usepackage{enumitem}

\begin{document}

\title{Efficient Audio-Visual Fusion for Video Classification}

\addauthor{Mahrukh Awan, Asmar Nadeem, Armin Mustafa}{{mahrukh.awan,asmar.nadeem, armin.mustafa}@surrey.ac.uk}{1}

\addinstitution{Centre for Vision, Speech and Signal Processing (CVSSP),\\ University of Surrey, UK}

\maketitle

\begin{figure}[t]
\includegraphics[width=\linewidth]{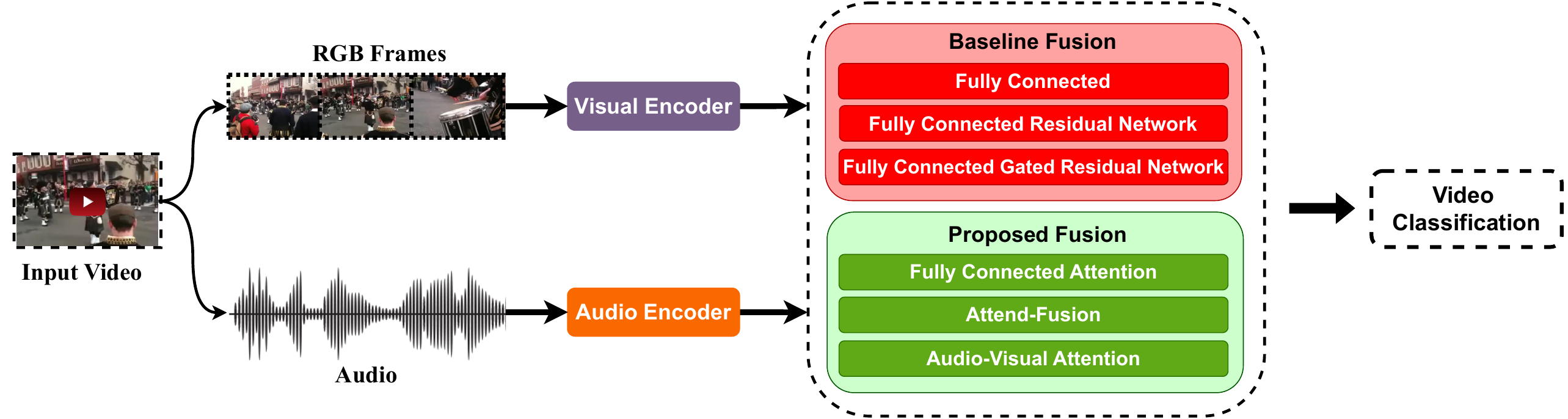}
\caption{Overview of our proposed audio-visual video classification framework on YouTube-8M dataset~\cite{abu2016youtube}, illustrating different fusion mechanisms of audio and visual modalities.}
\end{figure}

\noindent
We present Attend-Fusion, a novel and efficient approach for audio-visual fusion in video classification tasks. Our method addresses the challenge of exploiting both audio and visual modalities while maintaining a compact model architecture. Through extensive experiments on the YouTube-8M dataset~\cite{abu2016youtube}, we demonstrate that our Attend-Fusion achieves competitive performance with significantly reduced model complexity compared to larger baseline models.

The YouTube-8M dataset comprises millions of YouTube videos, each annotated with labels from a diverse vocabulary of 4,716 visual entities. Our approach focuses on effectively fusing audio and visual modalities to improve classification accuracy on this challenging multi-label dataset.

We compare our Attend-Fusion model with several baseline approaches, including Fully Connected (FC) networks. The FC Late Fusion model emerged as the best-performing baseline~\cite{bober2017cultivating}. Our Attend-Fusion model incorporates self-attention mechanisms~\cite{vaswani2017attention}, defined as:

\begin{equation}
\mathbf{X}_{att} = \text{softmax}\left(\frac{\mathbf{Q}\mathbf{K}^T}{\sqrt{d}}\right)\mathbf{V}
\end{equation}

where $\mathbf{Q}$, $\mathbf{K}$, and $\mathbf{V}$ are query, key, and value matrices derived from the input features, and $d$ is the feature dimension.

\begin{figure}[h]
\includegraphics[width=\linewidth]{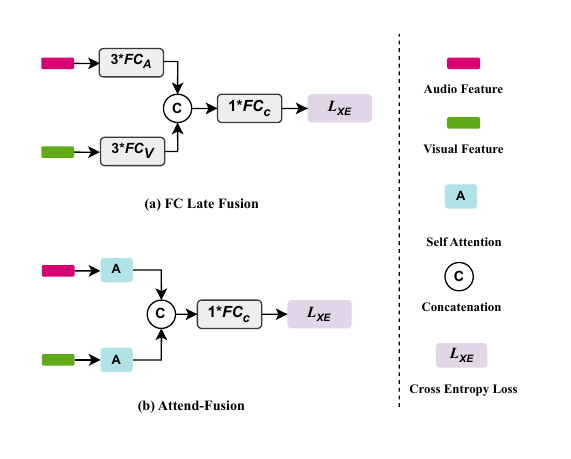}
\caption{Comparison of Fully-Connected (FC) Late Fusion (baseline) and Attend-Fusion architectures.\vspace{1mm}}
\end{figure}

\begin{table}[h]
\centering
\begin{tabular}{lcc}
\toprule
Metric & FC Late Fusion~\cite{bober2017cultivating} & Attend-Fusion \\
\midrule
GAP (\%) & 80.87 & 80.55 \\
F1 Score (\%) & 75.96 & 75.64 \\
Parameters (M) & 341 & 72 \\
\bottomrule
\end{tabular}
\caption{Comparison of FC Late Fusion and Attend-Fusion.\vspace{1mm}}
\label{tab:comparison}
\end{table}

Table \ref{tab:comparison} compares the performance and efficiency of FC Late Fusion and Attend-Fusion. Our model achieves comparable performance with significantly fewer parameters, demonstrating its efficiency.

The Attend-Fusion architecture processes audio and visual features separately through attention networks, which consist of fully connected layers and self-attention mechanisms. The attended features are then fused using a late fusion strategy, allowing the model to learn both modality-specific and cross-modal representations.

We conducted comprehensive ablation studies to investigate the contributions of different components. Removing the attention mechanism or using only a single modality led to significant drops in performance, confirming the value of our approach. The cross-entropy loss function used for multi-label classification is defined as:

\begin{equation}
\mathcal{L} = -\frac{1}{N}\sum_{i=1}^N\sum_{c=1}^C y_{i,c}\log(\hat{y}_{i,c}) + (1 - y_{i,c})\log(1 - \hat{y}_{i,c})
\end{equation}

where $N$ is the number of samples, $C$ is the number of classes, $y_{i,c}$ is the ground truth label, and $\hat{y}_{i,c}$ is the predicted probability.

\begin{figure}[h]
\includegraphics[width=\linewidth]{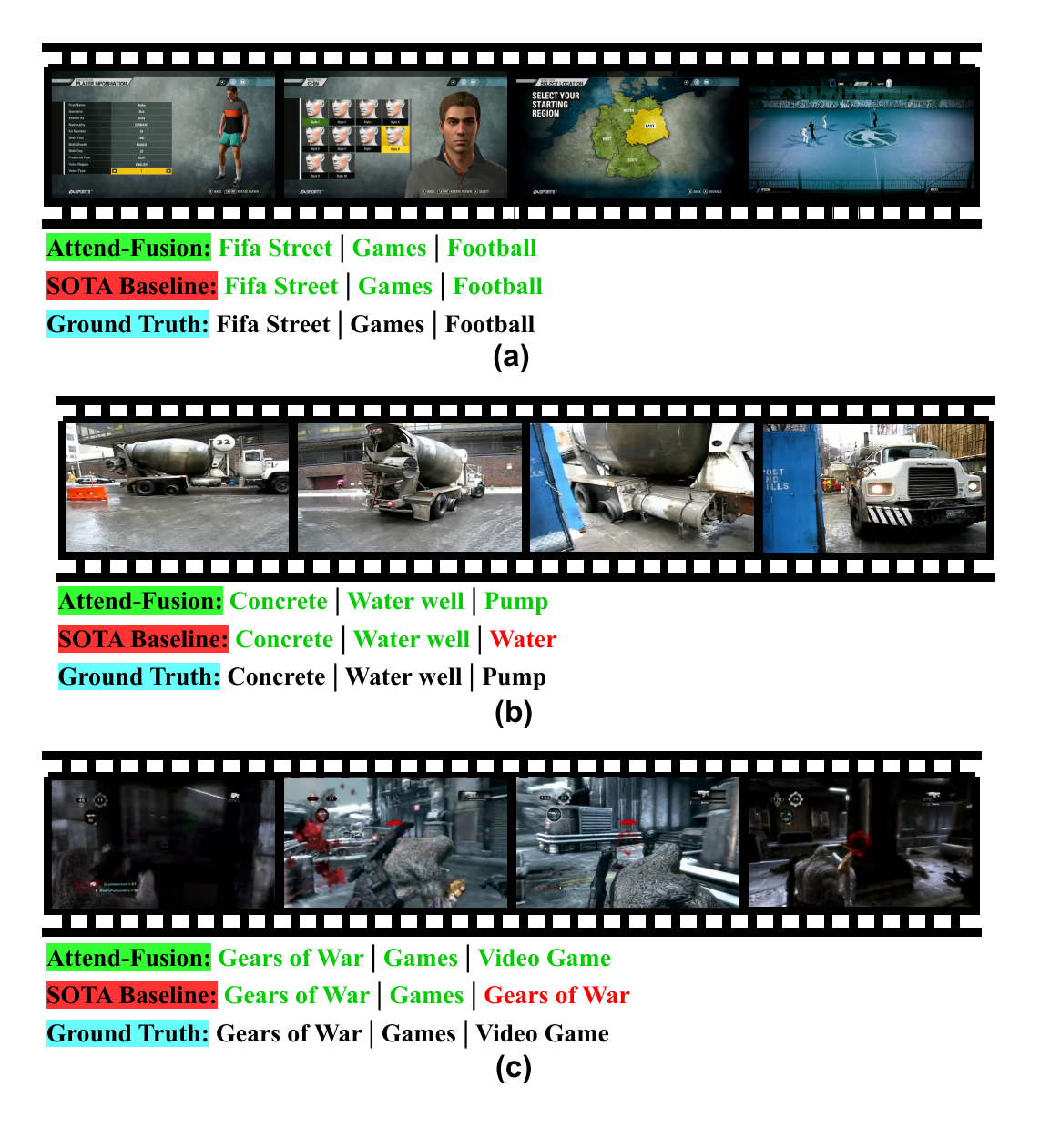}
\caption{Qualitative results comparing the top-3 predictions of Attend-Fusion, FC Late Fusion (SOTA baseline), and the ground truth labels on representative examples from the YouTube-8M dataset.}
\label{fig_3}
\end{figure}

Qualitative results in Figure~\ref{fig_3} on representative examples from the YouTube-8M dataset showcase Attend-Fusion's ability to accurately classify videos across various domains, including sports, music, gaming, and more. The model demonstrates superior performance in capturing fine-grained details and maintaining coherence in its predictions, highlighting the effectiveness of the attention mechanism in integrating audio and visual information.

Attend-Fusion's compact design and high performance make it well-suited for real-world applications where computational resources are limited, such as mobile devices or edge computing scenarios. Our work contributes to the ongoing efforts to develop more sustainable and deployable AI solutions for video understanding tasks, opening new possibilities for efficient video classification across various domains.

\end{document}